# Interpretable facial dynamics as behavioral and perceptual traces of deepfakes.

Timothy Joseph Murphy[1,2], Jennifer Cook[1], Hélio Clemente José Cuve[2]*

[1]School of Psychology, University of Birmingham, Edgbaston, Birmingham, B15 2TT, United Kingdom

[2]School of Psychology and Neuroscience, University of Bristol, 12a Priory Road, Bristol, BS8 1TU, United Kingdom

*Corresponding author.

**Email:**  helio.cuve@bristol.ac.uk

**Author Contributions (CRediT):**

TJM: Conceptualization, Data Curation, Formal analysis, Investigation, Methodology, Writing – Original, Visualization, Validation, Software

JLC: Funding acquisition, Writing – Review & Editing, Supervision

HCJC: Conceptualization, Data Curation, Methodology, Resources, Funding acquisition, Writing – Review & Editing, Supervision

**Competing Interest Statement:** The authors declare no competing interests.

**Abstract**

Deepfake detection research has largely converged on deep learning approaches that, despite strong benchmark performance, offer limited insight into what distinguishes real from manipulated facial behavior. This study presents an interpretable alternative grounded in bio-behavioral features of facial dynamics and evaluates how computational detection strategies relate to human perceptual judgments. We identify core low-dimensional patterns of facial movement, from which temporal features characterizing spatiotemporal structure were derived. Traditional machine learning classifiers trained on these features achieved modest but significant above-chance deepfake classification, driven by higher-order temporal irregularities that were more pronounced in manipulated than real facial dynamics. Notably, detection was substantially more accurate for videos containing emotive expressions than those without. An emotional valence classification analysis further indicated that emotive signals are systematically degraded in deepfakes, explaining the differential impact of emotive dynamics on detection. Furthermore, we provide an additional and often overlooked dimension of explainability by assessing the relationship between model decisions and human perceptual detection. Model and human judgments converged for emotive but diverged for non-emotive videos, and even where outputs aligned, underlying detection strategies differed. These findings demonstrate that face-swapped deepfakes carry a measurable behavioral fingerprint, most salient during emotional expression. Additionally, model-human comparisons suggest that interpretable computational features and human perception may offer complementary rather than redundant routes to detection.

**Significance Statement**
Deepfakes pose growing societal threats, yet current detection methods lack transparency and offer little insight into the facial behaviors that distinguish real from manipulated videos. We show that face-swapping disrupts measurable temporal facial dynamics, particularly during emotional expressions. This emotion-dependent effect is explained by the (current) face-swap process systematically degrading the structure of emotive facial dynamics. Model and human judgments converge for emotive content but rely on divergent underlying strategies, supporting complementary rather than redundant human-machine detection approaches.

**Introduction**

Deepfakes, synthetic media created by generative artificial intelligence (AI), have attracted growing concern due to their use in political manipulation, non-consensual explicit content, and identity fraud – enabled by increasingly accessible generation tools (1, 2). While deepfake detection research has expanded rapidly, the field has largely converged on deep learning (DL) approaches such as Convolutional Neural Networks and Recurrent Neural Networks, trained to identify spatial artifacts within frames (3) or temporal inconsistencies across sequences (4). Despite achieving state-of-the-art performance, these methods suffer from three critical limitations. First, they function as "black boxes", offering limited interpretability (understanding model mechanics) and explainability (articulating why a decision was made) – a significant problem for forensic and legal contexts (5). Second, they overlook the correspondence between model and human judgments, which is essential for human-in-the-loop detection workflows (6). Third, their computational demands limit practical deployment.

Bio-behavioral approaches - which focus on deviations from natural "statistics" of well-characterized human behavior - may provide explainable deepfake detection methods, but they commonly overlook highly transparent features such as facial dynamics. Recent bio-behavioral approaches have explored physiological signals such as remote photoplethysmography (7–9) and various facial dynamics including mouth movements (10), ear-mouth correlations (11), eyebrow dynamics (12), and expression patterns (13). Facial dynamics, systematically quantifiable through the Facial Action Coding System (FACS) (14) as discrete Action Units (AUs), provide a particularly promising basis for transparent detection. AUs correspond to specific facial muscle actions, preserving a clear link between features and observable behavior while offering a shared perceptual basis accessible to both computational models and human observers. However, previous research often relies on exhaustive statistical summaries and DL embeddings, in which the diagnostic information is distributed across high-dimensional latent representations, reintroducing the opacity they aim to resolve. Furthermore, few empirical studies have sought to isolate the particular features of facial dynamics that are diagnostically predictive of deepfakes. Consequently, transparent approaches, such as those capitalizing on facial dynamics, remain under-utilized.

Progress toward truly transparent facial dynamics-based detection has, however, been constrained by several factors. The temporal structure of AU time series is complex and high-dimensional, and critically, the diagnostic information is likely not carried by individual AUs in isolation but by coordinated patterns of activation across multiple AUs simultaneously. Identifying these patterns requires a method capable of recovering latent co-activation structure from the data, yielding components grounded in observable facial behavior rather than abstract statistical embeddings and preserving the interpretability that makes bio-behavioral approaches appealing in the first place.

However, given that facial dynamics can vary with different behavioral signals (e.g. speech, emotion expression), a key open question is how these different cues influence detection signal strength. Emotionally expressive videos are of particular interest, given that emotional behaviors are frequently targeted for deepfake manipulation. Yet, whether emotive facial dynamics influence deepfake detection has received little empirical attention. For instance, it is plausible that authenticity signals are stronger in video depicting emotive facial behavior. Emotional expressions involve tightly coordinated, multi-muscle activation patterns with characteristic temporal profiles (15, 16) that may be especially challenging for generative models to faithfully reproduce. Face-swap deepfakes are generated by encoder-decoder architectures and Generative Adversarial Networks, which learn to map the visual identity of one person onto another's facial movements. While these models achieve convincing identity transfer, the extent to which they preserve the fine-grained coordination underlying emotional expressions remains unclear (17, 18). If generative models systematically degrade the dynamics of emotional expressions, then emotive videos may carry a stronger diagnostic signal for detection. Indirect evidence supports this, including greater variability between real and manipulated recordings for expressive content (19), and work on static deepfake images demonstrating that synthetic faces exhibit more "average" statistical properties than real faces (20). However, no prior study has directly examined whether emotional expressions modulate deepfake detection performance, nor tested whether face-swapping selectively degrades emotion-related facial dynamics. Addressing this is therefore important for both understanding boundary conditions on detection and explaining the frequent manipulation of emotive cues by deepfakes.

A further factor limiting progress in deepfake detection concerns the lack of empirical work evaluating the correspondence between model predictions and human observer judgments. Indeed, human observers have been shown to display some detection biases that may, at a group level, provide a diagnostic cue for detection (6). Correspondence between model predictions and human observer judgments may offer useful data for optimizing detection tools: If model and human judgments systematically converge, this provides mutual validation and can support trust in model outputs. If they diverge, it may reveal complementary strengths that could be leveraged in integrated detection workflows (6). Despite this, prior bio-behavioral deepfake detection research has largely neglected this comparison, leaving open the question of whether humans are sensitive to the same diagnostic signals as computational models, which has implications for human-aligned AI systems (21).

The current study therefore addressed three objectives: (i) identify features of facial dynamics that are diagnostically predictive of deepfakes; (ii) characterize the influence of emotive facial behavior on detection signal strength; and (iii) evaluate the correspondence between model predictions and human observer judgments to assess whether humans are sensitive to the same facial dynamics as the model.

We focused on face-swapped deepfakes – in which a target's visual identity is replaced while theoretically preserving the original facial dynamics (22) – since they are among the most common manipulation types and are a significant impersonation threat (3). We adopted a person-generic approach, training models to detect statistical deviations from natural facial dynamics in general rather than learning identity-specific profiles, motivated by the aim of developing methods applicable to new individuals without extensive prior data on each individual (23). Specifically, we extracted facial AU intensity data from the Google DeepFakeDetection dataset (24), which contains sufficient variation in facial expressivity to enable emotion-stratified analysis without additional annotation. We then applied spatiotemporal dimensionality reduction using non-negative matrix factorization (NMF) (25) – an approach previously demonstrated to recover interpretable components of facial movement in related contexts (26) - to identify core low-dimensional patterns of coordinated facial movement. Using a theoretically informed time-series feature engineering approach (16, 26, 27), we derived metrics characterizing spatiotemporal structure. Consistent with our focus on interpretability, these features were used to train traditional machine learning classifiers, chosen for their transparency and computational efficiency. Finally, we validated and compared our model with results from a perceptual experiment in which human participants evaluated DeepFakeDetection stimuli based on isolated facial dynamics. Our results demonstrate that interpretable temporal features provide a diagnostic fingerprint for detection, particularly in emotive videos, and reveal significant but context-dependent convergence between model and human judgments.

**Results**
**Facial dynamics reveal low-dimensional interpretable patterns in real and deepfake videos.**
To create an interpretable basis for analysis, we applied NMF (25) to the facial AU data from the training set to identify core patterns of coordinated facial movement (see Materials and Methods). NMF is a dimensionality reduction technique that decomposes multivariate data into additive, non-negative components representing co-activation patterns across variables. NMF's non-negativity constraint yields parts-based representations that map naturally onto discrete muscle activations, preserving interpretability at the level of observable facial behavior. A three-component model was found to be optimal (see SI Appendix, S3).

Figure 1 shows the basis matrix and corresponding face maps for the three identified patterns. Component 1 captures a lower-face pattern dominated by coordinated cheek raising and lip corner activity (AU12, AU06, AU10, AU14). Component 2 reflects a distinct lower-face configuration involving chin raising, lip corner depression, and lip tightening (AU17, AU15, AU23). Component 3 isolates an upper-face pattern characterized by brow raising (AU01, AU02) and upper lid retraction (AU05). Notably, these components are additive: different combinations may contribute to the same expressive state, such that a given emotion is not necessarily indexed by a single component in isolation.

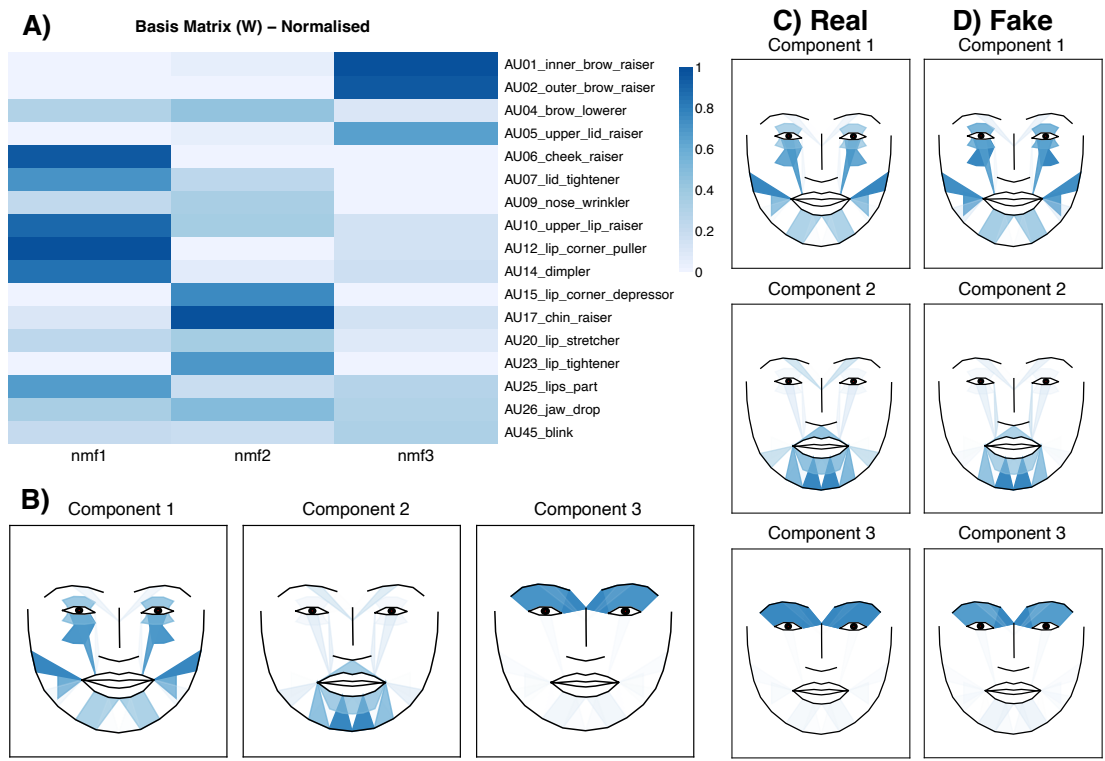

**Figure 1.** (A) Heatmap of the three-component basis matrix (W), showing AU loadings onto each component. Darker cells indicate stronger contributions of a given AU to that component. (B) Face maps visualizing the facial AU groups for Components 1-3 for both real and fake videos, (C) real videos only, and (D) fake videos only. Face maps were created using the py-feat library (28).

**Temporal (ir)regularities of facial dynamics provide a diagnostic feature-signal for deepfake detection.** Having identified core patterns of facial movement, the next step was to determine whether temporal features derived from these patterns could distinguish real from deepfake videos. For each NMF component, the AU with the highest loading was selected as a representative proxy, and a set of 111 temporal metrics characterizing spatiotemporal structure was extracted from each (see Materials and Methods). The features quantify the smoothness, rhythm, and variability of facial muscle activations over time, capturing higher-order kinematic properties (the predictability and regularity of movement velocity and acceleration) rather than movement magnitude itself (see SI-Table S2 for full behavioral definitions).

The Boruta feature selection algorithm (29) identified eight statistically important features (Figure 2A; SI-Table S2). The features broadly captured higher-order temporal properties of facial movement dynamics, including velocity rhythm, acceleration smoothness, distributional stability, and variability consistency. Features were distributed across lower-face (mouth, chin) and upper-face (brow) regions. The specific features identified should not be interpreted as a fixed or exhaustive set, as the exact configuration may vary given the stochastic nature of training and testing procedures.

Four classifiers were trained on these features using the training set (n=370) with 5-fold, 3-repeat cross-validation, and evaluated on a held-out test set (n=94). The classifiers were selected to span a range of model complexities while maintaining interpretability: Logistic Regression as a transparent linear baseline; C5.0 Boosted Decision Trees as a rule-based nonlinear comparator; Random Forest as an ensemble method providing native feature important estimates; and Support Vector Machine as a margin-based classifier offering a contrasting approach to decision boundary estimation. Full details of classifier training and evaluation procedures are provided in the classifier models and evaluation section of Materials and Methods. Model performance is summarized in Table 1. Non-linear models showed modest but statistically significant above-chance discrimination between real and fake videos. The Random Forest classifier achieved the best overall performance (Figure 2B), with a ROC-AUC of 0.694 [95% CI: 0.585, 0.804], accuracy of 66.0% [95% CI: 56.0%, 74.8%], and MCC of 0.322, indicating moderate diagnostic value. The distribution of predicted probabilities shows moderate separability between classes (Figure 2C). Confusion matrices for all classifiers are shown in Figure 2F; detailed performance analysis in SI-S5.

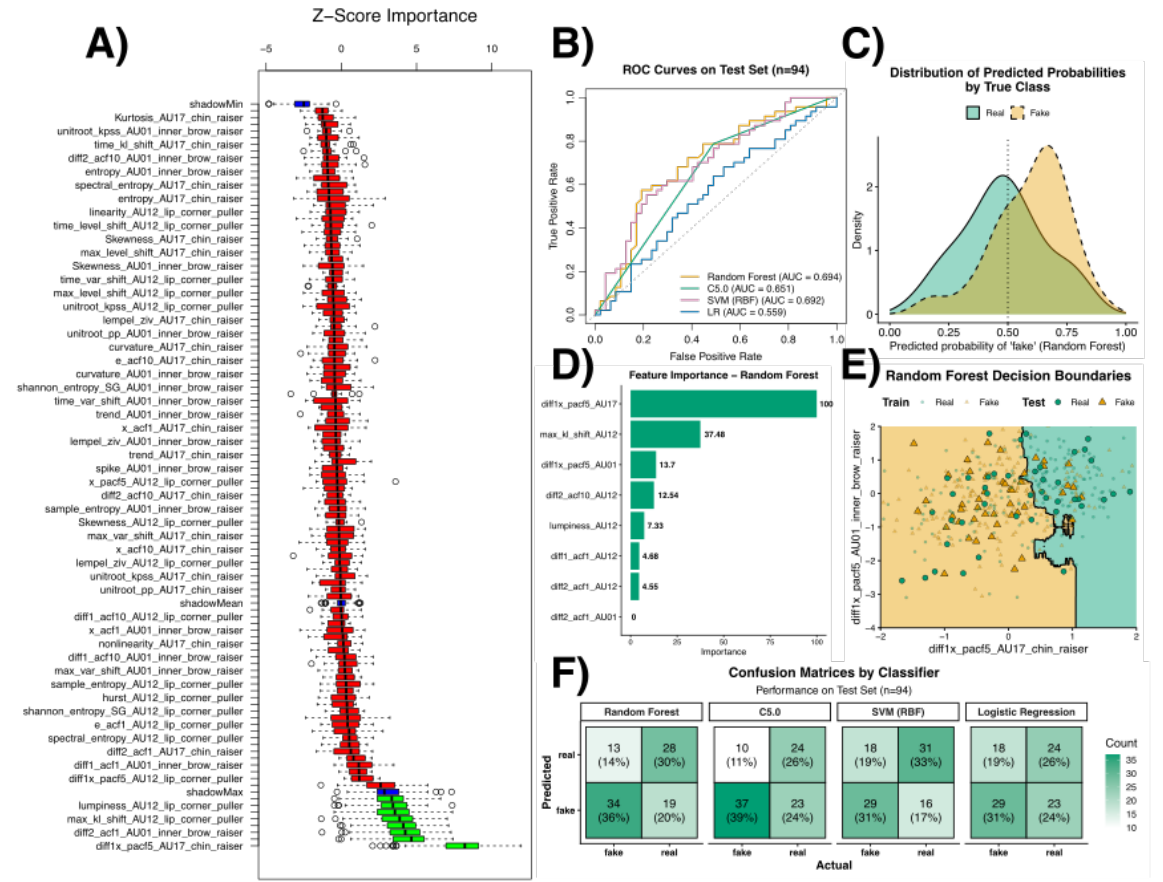

**Figure 2.** (A) Feature importance plot from the Boruta algorithm, showing the eight features confirmed as important (green). Feature names follow the convention *[transformation]_[metric]_[AU]* (e.g., *diff1* = velocity, *diff2* = acceleration; *acf* = autocorrelation, *pacf*

= partial autocorrelation; features without a prefix characterize distributional properties of the raw signal). (B) ROC curves for the four models evaluated on the test set. (C) Distribution of predicted probabilities for the Random Forest model, showing separability by true class. (D) Feature importance plot from the Random Forest model, ranked by Mean Decrease in Accuracy. (E) Decision boundary map for the Random Forest model, plotting the two most important autocorrelation features. (F) Confusion matrices for all four classifiers, showing predicted vs. actual counts on the test set.

**Table 1.** Deepfake classification performance.

| Model | ROC-AUC [95% CI] | Accuracy [95% CI] | MCC | $p$-value | Sensitivity [95% CI] | Specificity [95% CI] |
|---|---|---|---|---|---|---|
| Random Forest | **0.694** [0.585-0.804] | **66.0%** [56.0-74.8] | 0.322 | < .01 | 72.3% [62.5-80.3] | 59.6% [49.5-69.0] |
| C5.0B | **0.651** [0.558-0.744] | **64.9%** [54.8-73.8] | 0.310 | < .01 | 78.7% [69.4-85.8] | 51.1% [41.2-61.0] |
| SVM | **0.692** [0.585-0.799] | **63.8%** [53.3-73.5] | 0.277 | < .01 | 61.7% [51.6-70.9] | 66.0% [56.0-74.8] |
| Logistic Regression | **0.559** [0.441-0.676] | **56.4%** [46.3-66.0] | 0.128 | .128 | 61.7% [51.6-70.9] | 51.1% [41.2-61.0] |

Note: Models trained on eight important temporal features using 5-fold, 3-repeat cross-validation; evaluated on held-out test set (n=94). CIs for ROC-AUC calculated using DeLong's method; CIs for other proportion-based metrics using Wilson's score method. P-value reflects a one-proportion z-test against the no-information rate (0.5).

**Facial behavior type matters: emotive facial cues are more diagnostic than non-emotive facial cues.** To evaluate the influence of emotive cues on detection (Aim 2), we split the test set (n=94) into videos annotated as containing emotive cues (Emotion; n=58) and those without (No Emotion; n=36) and evaluated the Random Forest model on each subset. The distribution closely mirrored the training set (n=370; Emotion=222; No Emotion=148).

The model achieved 72.4% accuracy on Emotion videos, compared to 55.6% on No Emotion videos (Figure 3A). Performance on Emotion videos was significantly above-chance ($p$ < .001; MCC = 0.453), whereas performance on No Emotion videos was not ($p$ = .309; MCC = 0.134). However, direct comparison between subsets found that neither the accuracy difference (Fisher's exact $p$ = .119) nor the ROC-AUC difference (DeLong's test $p$ = .183) reached statistical significance. These comparisons should be interpreted cautiously given the small subset sizes and the conservative nature of the between-condition tests; whether this reflects a true absence of difference or insufficient power to detect one remains unclear. Accordingly, the magnitude of differences should be treated as indicative rather than definitive, though the contrast between significant and chance-level performance suggests a meaningful difference in detection signal strength.

A detailed breakdown (Table 2; Figure 3B) reveals that this performance gap is largely driven by a sharp drop in specificity for No Emotion videos: only 27.8% of real No Emotion videos were correctly classified (i.e., 72.2% false positive rate), despite high sensitivity (83.3%). In contrast, the Emotion subset showed balanced performance (sensitivity = 65.5%, specificity = 79.3%). This pattern suggests that when emotive cues are absent, the model encounters a weak signal and defaults to classifying videos as fake. This pattern is visible in the overlapping probability distributions (Figure 3C-D), ROC-AUC curves (Figure 3E-F), and less effective decision boundary (Figure 3H) for No Emotion videos, compared with the clearer class separation for the Emotion subset (Figure 3C, 3G). This pattern persists when emotion is matched between training and test splits, confirming that the effect is not driven by train-test imbalance in emotive content (see SI-S6).

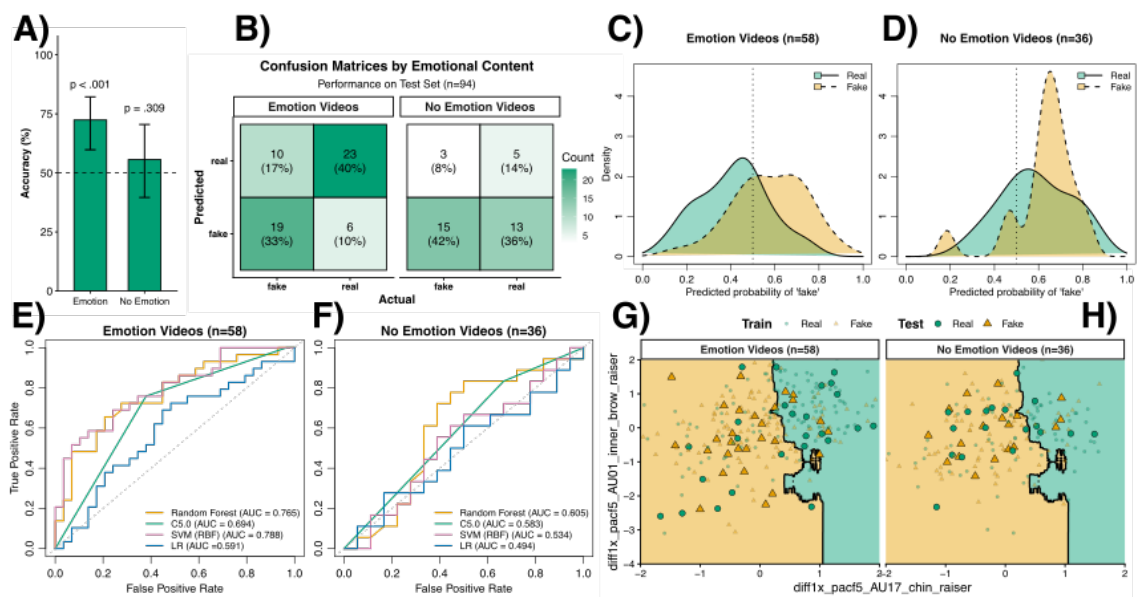

**Figure 3.** (A) Random Forest classification accuracy on Emotion (n=58) vs. No Emotion (n=36) held-out videos, with 95% CIs and the no-information rate (0.5) indicated by the dashed line. (B) Confusion matrices for each subset, showing predicted vs. actual counts for the Random Forest model. (C) and (D) show the distribution of predicted probabilities by true class (Random Forest), for Emotion and No Emotion subsets, respectively. (E) and (F) show ROC curves for the tested models for Emotion and No Emotion subsets, respectively. (G) and (H) show the corresponding decision boundary map for Emotion and No Emotion subsets, respectively.

**Table 2.** Performance of Random Forest classifier on Emotion and No Emotion subsets.

| Metric | Emotion Videos (n=58) | No Emotion Videos (n=36) |
|---|---|---|
| ROC-AUC [95% CI] | **0.765** [0.642-0.889] | **0.605** [0.407-0.803] |
| Accuracy [95% CI] | **72.4%** [59.8-82.2] | **55.6%** [39.6-70.5] |
| MCC | **0.453** | **0.134** |
| *P*-value | **< .001** | **.309** |
| Sensitivity [95% CI] | **65.5%** [52.7-76.4] | **83.3%** [68.1-92.1] |
| Specificity [95% CI] | **79.3%** [67.2-87.7] | **27.8%** [15.9-44.0] |

**Evidence for emotion-specific degradation of facial dynamics in deepfakes.** The improved detection accuracy for Emotion videos raises the question of whether the face-swap process selectively degrades emotion-related facial dynamics. To test this directly, we conducted an emotional valence classification analysis. A Random Forest classifier was trained exclusively on real videos to predict emotional valence (positive, neutral, negative) from temporal features extracted from the representative AU time series (see Materials and Methods). If emotive signals are faithfully preserved in deepfakes, the model should generalize similarly to real and fake test videos. Systematic degradation would instead be reflected in reduced valence classification performance on fake videos.

Model performance is presented in Table 3. The model classified valence significantly above-chance on real test videos (Accuracy = 59.6%, *p* = .006; note the no-information rate is 33% for three-way classification), but performance was lower and non-significant on fake test videos (Accuracy = 51.1%, *p* = .091). This reduction is consistent with the degradation hypothesis, suggesting that facial dynamics diagnostic of emotional valence are systematically altered in deepfakes.

Per-class analysis revealed emotion-specific effects. Sensitivity for positive emotions dropped from 81.2% to 62.5%, and for negative emotions from 50.0% to 25.0%, indicating that both were degraded by face-swapping, with negative expressions showing the most severe reduction. Neutral expression sensitivity showed the opposite pattern, increasing from 47.4% to 57.9%. This suggests that while the autoencoder degrades the complex, high-coordination patterns characteristic of emotive expressions, it produces more classifiable dynamics for neutral content. This is consistent with the bio-behavioral fingerprint of a deepfake being context-dependent, becoming most visible when the generative model must replicate the spatiotemporal demands of overt emotion. A sensitivity analysis using the full unbalanced dataset confirmed the same directional patterns (see SI-S7).

**Table 3.** Valence classification performance on real versus fake test subsets.

| Test Subset | ROC-AUC [95% CI] | Accuracy [95% CI] | MCC | *p*-value | SensPos [95% CI] | SensNeu [95% CI] | SensNeg [95% CI] |
|---|---|---|---|---|---|---|---|
| Real | **0.778** [0.636-0.920] | **0.596** [0.454-0.724] | 0.406 | 0.006 | **0.812** [0.679-0.898] | **0.474** [0.339-0.613] | **0.500** [0.363-0.637] |
| Fake | **0.657** [0.503-0.811] | **0.511** [0.373-0.648] | 0.255 | 0.091 | **0.625** [0.482-0.749] | **0.579** [0.437-0.709] | **0.250** [0.148-0.389] |
| *Δ(Real-Fake)* | *0.121* | *0.085* | *0.151* | - | *0.188* | *-0.105* | *0.250* |

Note: Random Forest trained exclusively on real videos with classes balanced via downsampling (n=72 per valence class). 95% CIs calculated using Wilson's score method for accuracy and sensitivity, and DeLong's method for ROC-AUC.

**Convergence and divergence in human and machine deepfake detection from facial dynamics: a perceptual dimension of explainability.** To evaluate the correspondence between model and human predictions (Aim 3), we compared Random Forest outputs with judgments from human observers (n=89) who evaluated deepfakes from Point Light Displays isolating facial dynamics (see Materials and Methods). Both agents were assessed on a shared held-out test set (n=40).

Average human accuracy was 50.5%, consistent with chance. The model achieved 55.0% on this subset (Figure 4A), which is lower than its overall 66.0% accuracy, likely reflecting the higher proportion of No Emotion videos (n=24 vs. Emotion n=16) in this subset. Despite these modest individual accuracies, a chi-square test revealed a significant relationship between model and human predictions ($\chi^2$ (1, N=40) = 4.43, *p* = .035, $\varphi$ = 0.33; Figure 4B). Critically, this correspondence was context-dependent: Emotion videos showed strong alignment (81.2% agreement; $\chi^2(1)$ = 4.06, *p* = .04, $\varphi$ = 0.504), whereas No Emotion videos showed none (54.2% agreement; $\chi^2(1)$ = 0.38; *p* = .537, $\varphi$ = 0.126; Table 4).

**Table 4.** Human-model correspondence by emotion content and viewing condition.

| Emotion | n | Model-Human Agreement | | | |
| --- | --- | --- | --- | --- | --- |
| | | Agreement | $\chi^2(1)$ | P-value | $\varphi$ |
| Yes | 16 | 81.2% | 4.064 | .044 | 0.504 |
| No | 24 | 54.2% | 0.381 | .537 | 0.126 |

However, further analysis revealed that this alignment reflects convergent output labels rather than a shared detection strategy. Human accuracy did not vary with the model's two most diagnostic features (Figure 4C-D), suggesting that humans were not sensitive to the same temporal properties that drove model predictions and indicating divergent underlying detection strategies despite agreement in output labels. A correctness correspondence analysis found no significant relationship between the specific videos correctly classified by each agent ($\chi^2(1)$ = 1.604, $p$ = .205, $\varphi$ = 0.2), suggesting that model and humans were not systematically fooled by the same stimuli.

To directly test whether temporal features could predict human judgments, we trained classifiers using two cross-validation strategies (see Materials and Methods). A separate Boruta analysis identified 34 features as important for predicting human responses, substantially more than the eight diagnostic for deepfake detection, with only four overlapping. The human-predictive features were more heavily weighted toward entropy-based measures and long-range temporal dependencies, reflecting sensitivity to unpredictability and extended temporal structure in facial movement. The model's features were dominated by higher-order autocorrelation of velocity and acceleration, capturing short-term repetitive patterns in movement dynamics. This distinction suggests that humans and the model may extract fundamentally different aspects of facial motion when making authenticity judgments.

Leave-One-Participant-Out (LOPO) cross-validation achieved 59.7% accuracy, indicating modest prediction when generalizing to unseen observers (Figure 4, Row 2). However, Leave-One-Stimulus-Out (LOSO) cross-validation achieved only 56.9% accuracy. Both models exhibited a specificity bias when generalizing to new videos (Figure 4E-H), indicating that human judgments are either highly stimulus-specific, not fully captured by AU-based temporal features, or substantially variable across observers. When LOPO and LOSO-predicted human judgments were compared with the detection model's predictions on the broader test set (n=94) and full dataset (n=464), no significant alignment was found (Figure 4F, I-L; see SI Appendix, Section 10.3 for full methodology and results).

Together, these findings demonstrate that while model and humans arrive at similar final judgments for emotive videos, they rely on divergent detection strategies, suggesting complementary rather than redundant routes to detection and potential for integration in human-in-the-loop workflows.

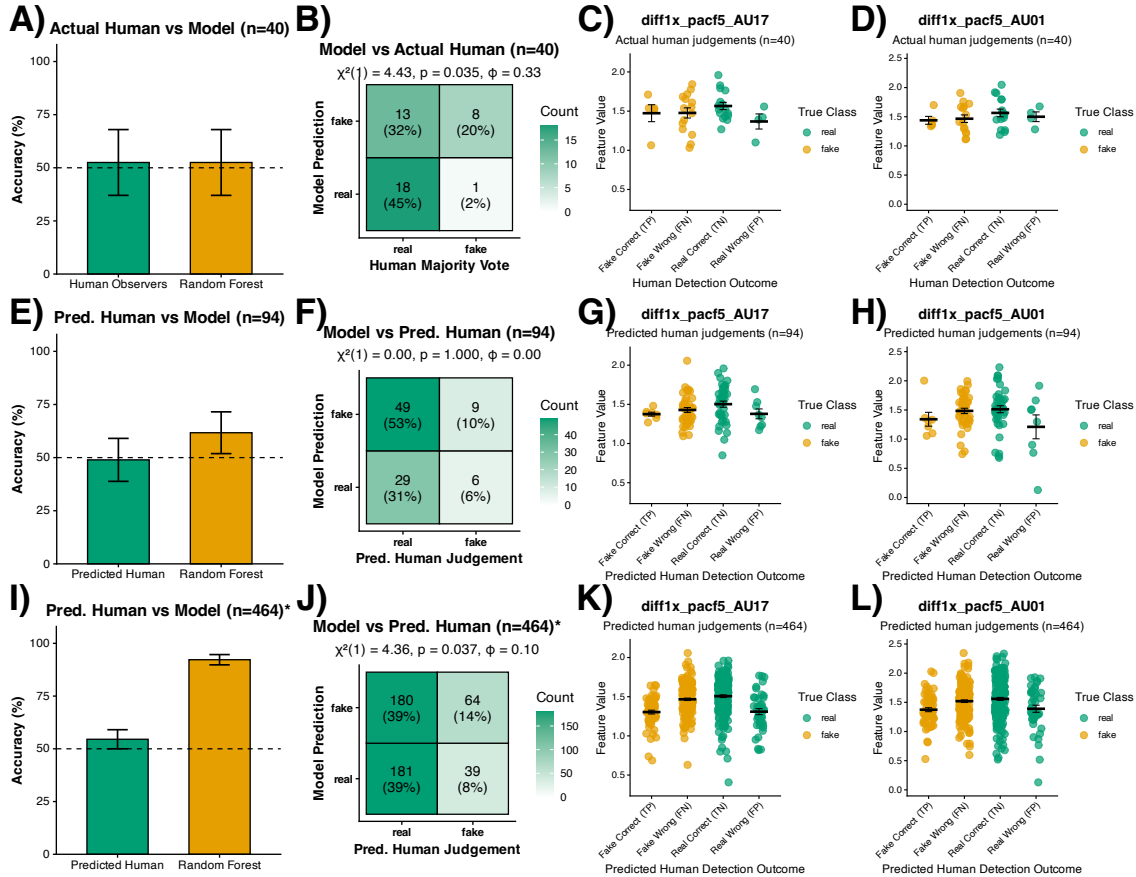

**Figure 4.** Human and machine detection strategies across actual and predicted human judgments. Rows 1-3 show human judgments (n=40 test set), LOPO-predicted human judgments (n=94), and predictions extended to full dataset (n=464), respectively. (A, E, I) Detection accuracy for human observers and the Random Forest model; error bars indicate 95% CIs. Model accuracy for the n=464 set (I) reflects performance on the combined dataset including training data and is not a valid generalization estimate; the Random Forest bar is shown for reference only. (B) Agreement matrix showing significant human-model alignment ($\chi^2(1) = 4.43$, $p = .035$, $\varphi = 0.33$); (F) no significant alignment ($\chi^2(1) = 0.00$, $p = 1.00$, $\varphi = 0.00$); (J) agreement matrix ($\chi^2(1) = 4.36$, $p = .037$, $\varphi = 0.10$). (C-D, G-H, K-L) Feature distributions across detection outcome categories; crossbars indicate means ± SE. Row 3 includes training data for visualization purposes only.

**Discussion**

Despite growing interest in deepfake detection, fundamental questions remain about what distinguishes real from manipulated facial behavior and how automated decisions relate to human perception. Dominant deep learning approaches often achieve strong benchmark performance but offer limited interpretability (5, 30, 31), which is particularly problematic given increasing interest in human-machine detection workflows where understanding convergence and divergence between agents is essential (6). The present study addressed these gaps by grounding deepfake detection in interpretable features of temporal facial dynamics combined with traditional and more interpretable machine learning methods.

Three key findings emerged. First, a small number of structured facial dynamic patterns, identified through non-negative matrix factorization, provided a principled basis for extracting interpretable features for detection. These features carried a subtle but significant diagnostic signal, driven by higher-order temporal irregularities (deviations in the predictability and smoothness of facial dynamics) rather than movement magnitude. Second, this signal was behavior-dependent: detection was substantially more reliable for emotive facial behavior, and a valence classification analysis provided direct evidence that face-swapping systematically degrades emotion-related dynamics. Third, while model and human judgments converged for emotive videos, deeper analysis revealed strategic divergence. Human accuracy did not covary with the model's most diagnostic features, and humans and models were fooled by different stimuli, indicating complementary rather than redundant detection routes.

The finding that face-swapping disrupts spatiotemporal structure more strongly in emotive videos is consistent with emotive facial behavior containing stronger temporal regularities that are more vulnerable to degradation by generative models. The face-swap process aims to map between identities while preserving plausible motion, yet our results suggest it nonetheless disrupts spatiotemporal coordination, particularly for the complex, multi-muscle patterns characteristic of emotional expression. This indicates that the bio-behavioral fingerprint of a deepfake is context-dependent, becoming most detectable when the generative model is tasked with reproducing the demands of overt emotion. The reasons underlying this asymmetry remain unclear; it may reflect biases in training data for generation models (32, 33), or inherent limitations in how autoencoders represent expressive behavior, potentially yielding more stereotyped outputs (17, 18). Future work should examine whether models optimized for motion transfer (e.g., facial reenactment) or fully synthesized approaches (e.g., diffusion-based models) exhibit distinct patterns of disruption, clarifying whether this asymmetry is a general property of generative manipulation or specific to autoencoder-based face-swapping.

A potential counterargument is that the emotion-dependent performance reflects variation in speech-related facial movement across subsets rather than genuine degradation of emotive dynamics. Videos involving overt speech produce characteristic mouth dynamics that could drive detection independently of emotional content. However, both the Emotion and No Emotion subsets of the Google DeepFakeDetection dataset contain speech, meaning speech-related dynamics are present across conditions and are unlikely to account for the performance difference between them. Any residual differences in facial movement between subsets more plausibly reflect genuine differences in expressive behavior rather than a speech artifact.

The absolute performance of our approach is modest compared to state-of-the-art methods. However, this reflects a deliberate trade-off: our pipeline relies exclusively on AU dynamics, whereas many deep learning methods derive features from dense raw video data including spatial cues and optical flow (24, 34, 35). This restriction substantially limits available information but preserves a clear link between features and observable facial actions. Viewed in this context, the observed above-chance performance indicates a genuine behavioral signal, and the principles learned here may be applicable to probing explainability in higher-performing but less transparent models, particularly given emerging evidence that deepfakes struggle to reproduce emotional behavioral dynamics (18–20).

The comparison with human observers provides an additional perceptual dimension of explainability. By constraining observers to the same class of perceptual information available to the model, specifically the temporal dynamics of facial movement isolated via point-light displays, we were able to make comparisons between detection strategies that are more controlled than typical correlation comparisons. If both humans and the model have access to equivalent information and still diverge in their strategies, this provides stronger evidence that the divergence reflects genuine differences in the underlying processing strategies of each, rather than differences in available information. For emotive videos, strong alignment in final judgments suggests both models and humans are sensitive to spatiotemporal signals. Yet strategic divergence, revealed through correctness correspondence analysis and feature-response comparisons, indicates that humans may rely on qualitatively different perceptual cues. Understanding both convergence and divergence can inform integrated human-machine systems where complementary strengths are leveraged to improve performance (6).

Several limitations should be acknowledged. First, our analysis was confined to a single generation method (Faceswap autoencoder) (1) and dataset (Google DFD) (24); whether similar signatures emerge for alternative methods remains unclear. Second, the statistical comparison of classification accuracy between Emotion and No Emotion subsets was underpowered; post hoc power analysis on the two-proportions test indicated only 38.5% power to detect the observed effect size (Cohen's $h$ = 0.35, $n_1$ = 58, $n_2$ = 36, $\alpha$ = .05), well below the conventional 80% threshold (36). A sample of approximately 130 videos per group would be required to achieve adequate power for this inferential comparison. The coarse emotion annotations also mean that "No Emotion" videos may contain subtle emotive behaviors, attenuating observed effects; this reflects a broader limitation of the field, as most deepfake datasets do not label videos by emotional content or systematically vary the type of facial behavior they contain, making emotion-stratified analyses impossible without additional annotation. Replication with larger datasets and finer-grained annotations is therefore warranted. Third, while AUs capture rich expressive behavior, they are less optimized for speech-related dynamics; future work should compare representations optimized for different types of facial behavior.

**Conclusion.** Face-swapped deepfakes introduce measurable alterations to facial dynamics that constitute a behavior-dependent bio-behavioral fingerprint, most salient during emotive expression. Human and model judgments converge for emotive but not non-emotive content, yet rely on divergent underlying strategies, supporting complementary rather than redundant detection routes. These findings contribute to a more interpretable account of deepfake detection grounded in natural facial behavior, with implications for human-machine systems where understanding when and why decisions align is essential.

**Materials and Methods**

**Dataset.** Videos were sourced from the Google DeepFakeDetection (DFD) dataset, a subset of FaceForensics++ (24), containing face-swapped deepfakes created using the DeepFakes autoencoder method (1). All 363 unique real videos were selected, and for each, one corresponding fake video was randomly chosen from available deepfake versions, yielding 726 videos (363 real/fake pairs). Videos featured diverse actor identities, facial expressions, lighting conditions, and backgrounds, recorded at 24 frames per second (fps).

To address Aim 2, videos were categorized based on the presence of overt emotional expressions derived from filename labels (angry, disgust, laughing, serious, happy). Videos labelled as "serious" or without emotion labels were categorized as "No Emotion". Valence categories were assigned as positive (laughing, happy), negative (angry, disgust), or neutral (serious, unlabeled).

**Preprocessing and facial action unit extraction.** Facial Action Unit (AU) intensities were extracted from each frame using OpenFace 2.0 (37), yielding continuous intensity values for 17 AUs. To reduce frame-level noise while preserving rapid facial dynamics, a 4-frame left-aligned rolling mean was applied to each AU time series independently, with endpoint extension to prevent boundary artifacts. AU intensities were z-score normalized per video, then shifted by subtracting the global minimum z-score from the training set to satisfy the non-negativity constraint for NMF (25). All videos were truncated to 241 frames (10 seconds at 24 fps) to control for variable video lengths and maintain consistency with stimuli used in the human participant experiment. Videos underwent multi-stage filtering to ensure data quality (see SI Appendix, Section 1).

**Dimensionality reduction and feature extraction.** A multi-stage feature engineering process was employed, consisting of spatiotemporal dimensionality reduction to identify core facial movement patterns, followed by time-series feature extraction from representative AUs (Figure 5).

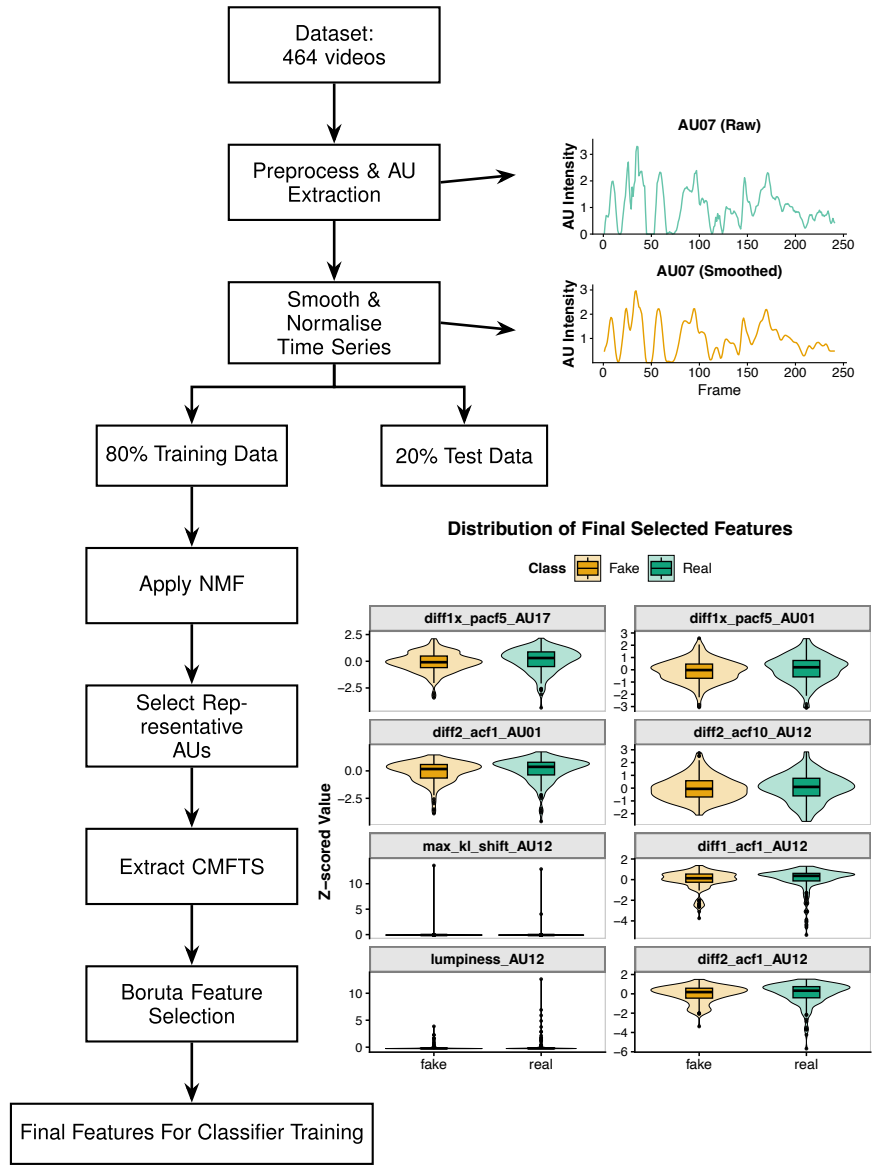

**Figure 5.** The pipeline extracts and preprocesses facial action units, applies NMF-guided feature selection, and identifies eight temporal features that distinguish real from fake videos. Distributions show standardized feature values for the final selected features, with real (green) and fake (orange) showing distinct patterns.

In step 1, the 17 AU dimensions were reduced into interpretable, coordinated movement patterns using NMF applied to the training data. NMF approximates the original data matrix as a weighted combination of additive, non-negative components, yielding a basis matrix (W) specifying AU contributions to each pattern and an activation matrix (H) describing component expression over time (25). A three-component model was selected based on reconstruction error across ranks 2-10 (see SI Appendix, Section 2), alignment with prior work on facial dynamics (26), and interpretability. The model explained approximately half of the variance ($R^2 = 0.42$).

In step 2, rather than extracting features from the NMF activation time series (subject to variance loss), we used the basis matrix as an interpretable guide: the AU with the highest loading onto each component was selected as a representative proxy for that core movement pattern. This grounded subsequent analysis in specific facial actions while retaining the holistic structure identified by NMF. For each selected AU, temporal features were extracted using the Complexity Measures and Features for Time Series (CMFTS) R package (27), which quantifies time-series dynamics using interpretable, mathematically defined metrics (e.g., entropy, autocorrelation, Lempel-Ziv complexity). This framework is consistent with features shown to encode emotion (16, 38) and identity cues (39, 40) in facial motion, and enables the use of traditional machine learning classifiers that would otherwise be unsuitable for raw time-series data.

The Boruta feature selection algorithm (29) was then applied to identify a parsimonious and predictive subset. Features confirmed as "Important" were retained for classifier training (see SI-Table S2).

**Classifier models and evaluation.** Four classifiers were used: Random Forest, C5.0 Boosted Decision Trees, Support Vector Machine with radial basis function kernel, and Logistic Regression. Non-linear models were chosen for their ability to capture complex relationships. Logistic Regression was included as a linear benchmark. All models were trained using the *caret* package in R (version 4.4.1) on the training data (n=370) with 5-fold, 3-repeat cross-validation, and evaluated on the held-out test set (n=94).

Performance was assessed using ROC-AUC, Accuracy, Matthews Correlation Coefficient (MCC), Sensitivity, and Specificity (41, 42). Statistical significance was determined via the one-proportion z-tests against the no-information rate (43). 95% CIs for ROC-AUC were calculated using DeLong's method (44); CIs for proportion-based metrics used the Wilson score method (45).

To address Aim 2, the test was partitioned into Emotion (n=58) and No Emotion (n=36) subsets, with differences assessed using Fisher's exact test (accuracy) and DeLong's test (ROC-AUC). A separate valence classification analysis was conducted, in which a Random Forest model was trained exclusively on real videos to predict emotional valence (positive, neutral, negative), and then evaluated on real and fake test subsets independently. Valence was selected for two reasons: it provides finer-grained differentiation of emotional behavior than a binary emotion/no-emotion distinction, and it yields a more balanced class distribution than discrete emotion categories (e.g., happy, angry, disgust), which could not be used due to unequal sample sizes across the dataset.

**Human participant experiment and convergence with model.** Data were collected from 89 participants (70 female, 17 male, 1 non-binary, 1 undisclosed; mean age = 19.73, SD = 1.79) recruited from the university community for course credits as part of a larger study on facial dynamics and perception.

Participants judged authenticity (real or fake) of Point Light Display versions of the deepfake stimuli, a format that isolates spatiotemporal facial dynamics by representing landmarks as moving dots on a black background, removing all textural information and visual artifacts (16). After each 10-second video, participants provided continuous authenticity ratings (0-100 scale; 0 = Real, 100 = Fake) via a visual slider. The 40 experimental videos (20 real, 20 fake) were presented in randomized order, preceded by four practice trials.

Continuous ratings were binarized at a threshold of $\geq 50$ and aggregated via majority vote per video. Human-model correspondence was assessed using chi-square tests of independence with phi coefficient ($\varphi$), performed on the full video set (n=40) and stratified by emotion content.

Additional analyses examined: (i) correctness correspondence (whether the same videos fooled both agents; chi-square on error patterns); (ii) whether human accuracy varied with the model's most diagnostic features; and (iii) whether AU-based features could predict individual human responses via Leave-One-Participant-Out and Leave-One-Stimulus-Out cross-validation. A separate Boruta analysis was conducted with human binary judgments as the prediction target to identify human-specific features. Predicted human judgments were extended to the broader test set (n=94) and full dataset (n=464) to assess robustness (see SI Appendix, Section 10.3).

**Ethics statement.** This study was approved by the University of Bristol Research Ethics Committee (ethics code: 21918). All participants provided informed consent. The study used publicly available data from Google DeepFakeDetection dataset (24).

**Acknowledgments**
The authors thank Chaeyeon Lim for data collection in the human participant experiment.

**Funding Information**
This work was supported by ESRC CENTRE-UB grant ES/Y002148/1. This work was partly funded by an Experimental Psychology Society Postdoctoral Fellowship awarded to HCJC. JLC was supported by the European Union's Horizon 2020 Research and Innovation Programme under ERC-2017-StG Grant Agreement No. 757583 (Brain2Bee; Jennifer Cook PI).

**Data and Code Availability**
The Google DeepFakeDetection dataset is publicly available as part of FaceForensics++ (24) at https://github.com/ondyari/FaceForensics. Supporting data for this study are openly available in the Open Science Framework at https://doi.org/10.17605/OSF.IO/76AGJ (46). Supporting code and analyses are archived on Zenodo at https://doi.org/10.5281/zenodo.19632073 (47), with the development repository hosted at https://github.com/tj-murphy/paper-deepfake-facial-dynamics.

**Supporting Information for**
Interpretable facial dynamics as behavioral and perceptual traces of deepfakes.

Timothy Joseph Murphy[1,2], Jennifer Cook[1], Hélio Clemente José Cuve[2]*

[1]School of Psychology, University of Birmingham, Edgbaston, Birmingham, B15 2TT, United Kingdom

[2]School of Psychology and Neuroscience, University of Bristol, 12a Priory Road, Bristol, BS8 1TU, United Kingdom

*Corresponding author.

**Email:** helio.cuve@bristol.ac.uk

**This PDF file includes:**

    Supporting text
    Figures S1 to S3
    Tables S1 to S6
    SI References

**Other supporting materials for this manuscript include the following:**

    None

**Supporting Information Text**

**S1. Dataset preprocessing and filtering.**

**S1.1 Dataset construction and pairing.** A custom Python script (*select_fakes.py*) paired each unique real video with one randomly selected corresponding fake version. OpenFace 2.0 (1) extraction was automated using a custom script (*process_openface.py*) with the *-aus* flag to limit output to AU data only. Videos that failed to process were automatically excluded.

**S1.2 Data cleaning and quality control.** After extraction, videos without corresponding real/fake pairs were removed, resulting in 358 matched pairs (716 videos). Binary AU presence columns (suffixed with _c) were excluded, retaining only intensity columns (suffixed with _r).

Videos containing fewer than 241 frames were excluded along with their corresponding pairs, removing 14 pairs (28 videos). The 40 videos reserved for the human participant study were treated as a held-out test set. Of these, 20 (19 real, 1 fake) overlapped with the current dataset due to the random pairing procedure described above, and were removed prior to further preprocessing.

OpenFace provides per-frame confidence scores (0-1) and a binary success indicator for tracking. The distribution showed that 88.77% of frames had confidence > 0.9 and 96.40% had confidence > 0.8 (median confidence = 0.971, mean success = 1.0). Initial quality thresholds of 0.88 (confidence) and 0.99 (success), derived from boxplot non-outlier minima, would have excluded 246 videos (~38% of the dataset), raising concerns about limiting generalizability to static, easily tracked scenes. To balance data quality with diversity, thresholds were reduced by 0.05 to 0.83 (confidence) and 0.94 (success), excluding 62 pairs (124 videos, ~19%).

Analysis of excluded videos revealed that certain scene types were disproportionately affected: all *walk_down_hall_angry* videos (n=28) were removed, as were 18 of 25 *walking_down_street_outside_angry* videos. These scenes typically featured subjects starting far from the camera and walking closer, resulting in poor landmark detection in early frames. Conversely, scenes with static, frontal subjects (e.g., *podium_speech_happy*, *talking_angry_couch*) showed no losses.

A random representative subset (n=82, calculated using a finite population sampling formula with 95% confidence and 10% margin of error) was visually inspected. This revealed that some videos contained multiple visible faces, causing inconsistent tracking. All videos from two multi-subject scene types were removed: *walking_and_outside_surprised* (n=13) and *walking_down_indoor_hall_disgust* (n=15). Four additional videos flagged for tracking errors were removed with their pairs.

The final dataset comprised 464 videos (232 real/fake pairs), each containing 241 frames of high-quality AU data.

**S1.3 Train-Test Split.** The dataset was split 80:20 using stratified sampling that preserved real/fake pairs. Valence distributions were: training (144 positive, 154 neutral, 72 negative) and test (32 positive, 38 neutral, 24 negative). Normalization parameters were learned exclusively from the training set to prevent data leakage. AUs with zero standard deviation were assigned SD = 1 to avoid division by zero, with normalized values set to zero.

**S2. NMF model selection.**

**S2.1 Implementation.** Non-negative matrix factorization (NMF) (2) was implemented using the RcppML R package (3), which provides an alternating least squares algorithm optimized for speed via the Eigen C++ library and parallelized using OpenMP multithreading. We initially attempted to use the NMF R package (4), which offers built-in rank selection utilities (e.g., cophenetic correlation), but this proved computationally prohibitive due to impractically long runtimes. RcppML offered substantially faster computation while producing equivalent factorizations. Key implementation differences include diagonalized scaling of factors (where component weights are stored in a separate diagonal vector, *d*, rather than absorbed into the basis or activation matrices), fast convergence criteria based on inter-iteration correlation, and native support for parallelization.

Because RcppML stores component scaling separately, the standard reconstruction is computed as $V \approx W \cdot d \cdot H$. For all downstream analyses, $d$ was manually applied to scale the appropriate matrices.

**S2.2 Rank selection.** Ranks $k=2$ to $k=10$ were systematically evaluated by calculating Mean Squared Error (MSE) between original and reconstructed matrices for each rank. As shown in Figure S1, MSE decreased monotonically with no sharp elbow. We selected $k=3$ based on: (i) diminishing returns in reconstruction error beyond $k=3$, (ii) alignment with prior work identifying three core spatiotemporal patterns in emotive facial expressions (5), and (iii) the goal of maintaining interpretability, as three components can be meaningfully visualized and interpreted.

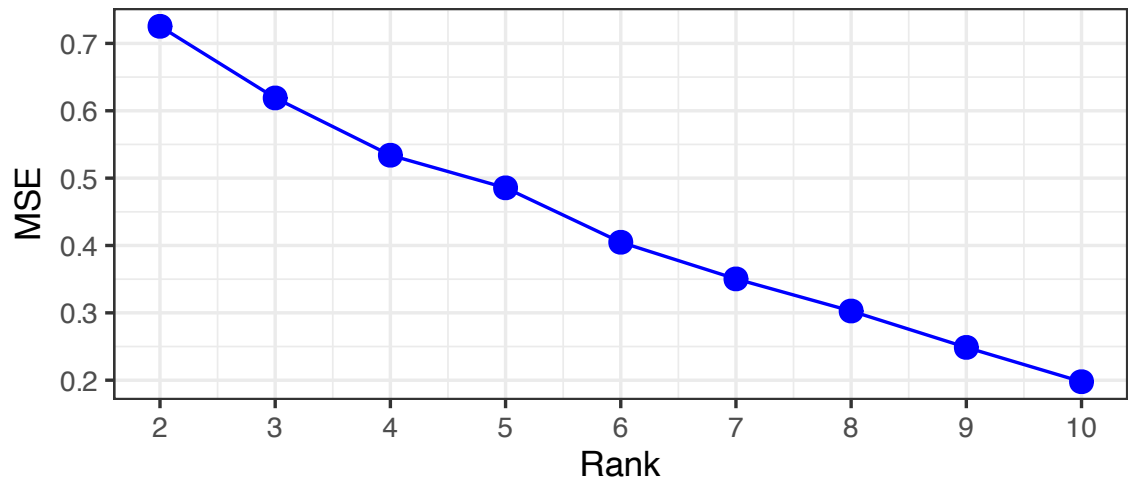

**Fig. S1.** MSE vs. rank plot for determining optimal NMF rank.

**S2.3 Reconstruction quality.** To assess how well the three-component NMF model captured the dynamics of individual AUs, per-AU reconstruction $R^2$ values were calculated on the test split (n=94) and the held-out set (n=40; Table S1). Reconstruction quality varied substantially across AUs. Upper-face and brow-related AUs were generally well-reconstructed (e.g., AU01: $R^2 = 0.744$; AU02: $R^2 = 0.774$), as were mouth-related AUs (AU06: $R^2 = 0.788$; AU12: $R^2 = 0.708$). In contrast, AUs associated with more subtle or less frequently activated movements showed substantially lower reconstruction (e.g., AU04: $R^2 = 0.007$; AU45: $R^2 = -0.019$). The overall $R^2$ of 0.42 on the training set and 0.45 on the test set indicates that the model captured meaningful coordinated patterns while leaving substantial AU-specific variance unexplained. This uneven reconstruction quality motivated the representative AU pipeline.

**Table S1.** Per-AU reconstruction $R^2$ for the test split (n=94) and held-out set (n=40).

| AU | $R^2$ (Test Set; n=94) | $R^2$ (Held-Out Set; n=40) |
|---|---|---|
| AU06 (Cheek Raiser) | 0.788 | 0.606 |
| AU01 (Inner Brow Raiser) | 0.744 | 0.686 |
| AU02 (Outer Brow Raiser) | 0.774 | 0.700 |
| AU12 (Lip Corner Puller) | 0.708 | 0.443 |
| AU17 (Chin Raiser) | 0.541 | 0.555 |
| AU14 (Dimpler) | 0.519 | 0.164 |
| AU10 (Upper Lip Raiser) | 0.388 | 0.267 |
| AU07 (Lid Tightener) | 0.385 | 0.183 |
| AU05 (Upper Lid Raiser) | 0.284 | 0.303 |
| AU25 (Lips Part) | 0.270 | 0.120 |
| AU23 (Lip Tightener) | 0.245 | 0.185 |
| AU15 (Lip Corner Depressor) | 0.242 | 0.350 |
| AU26 (Jaw Drop) | 0.207 | 0.096 |
| AU20 (Lip Stretcher) | 0.095 | 0.076 |
| AU09 (Nose Wrinkler) | 0.056 | 0.060 |
| AU04 (Brow Lowerer) | 0.007 | -0.053 |
| AU45 (Blink) | -0.019 | 0.102 |

### S3. Feature engineering and selection.

**S3.1 CMFTS feature extraction.** Temporal features were extracted from the three representative AU time series (AU01, AU12, AU17) using the Complexity Measures and Features for Time Series (CMFTS) R package (6). For each AU, 37 base features were calculated, comprising complexity measures (e.g., entropy measures, forbidden patterns) and time series features (e.g., autocorrelation, trend, stationarity). With three representative AUs selected, the total feature set comprised 111 features (37 features × 3 AUs), each suffixed by the corresponding AU identifier (e.g., *diff1_acf1_AU01*).

Several features were computed on first-order (*diff1*) and second-order (*diff2*) differences of the AU time series, representing the frame-to-frame velocity and acceleration of AU activation respectively. Computing autocorrelation and partial autocorrelation on these differenced series captures temporal regularity at progressively higher orders of kinematic complexity.

After extraction, features with zero standard deviation across the training set were removed. Permutation entropy features were excluded due to missing values. Infinite values in Shannon entropy features were imputed using the missForest algorithm (7), with an out-of-bag normalized root mean squared error of 0.004.

**S3.2 Boruta feature selection.** The Boruta feature selection algorithm (8) was applied with default parameters, using the real/fake label as the prediction target. Eight features were confirmed as important, with no tentative features remaining after the *TentativeRoughFix* procedure. Features are described in Table S2.

### S4. NMF component classifier results.

Prior to developing the representative AU pipeline reported in the main text, we attempted to classify real and fake videos using temporal features derived directly from the NMF component activation time series (H matrix). CMFTS features were extracted from the three component activations, producing 333 total features (111 per component).

This approach yielded weak and unstable performance. Boruta failed to identify any features as statistically important. Recursive Feature Elimination (9) identified a subset of features, but results were inconsistent across configurations and model types. The best-performing Random Forest achieved approximately 58-63% accuracy depending on the feature set and balancing strategy, but performance was not consistently significant. C5.0 Boosted Decision Trees frequently collapsed to predicting a single class, and performance on the held-out set (n=40) was generally at or below chance.

This failure is attributable to information loss during NMF decomposition. The uneven reconstruction quality across AUs (Table S2) means that features derived from NMF activations captured broad movement structure but not the fine-grained temporal properties diagnostic of manipulation. This motivated the representative AU pipeline.

**Table S2.** Boruta-selected features (n=8) with interpretative descriptions.

| Feature Identifier | AU | Base Feature | Transformation | Behavioral interpretation |
|---|---|---|---|---|
| diff1x_pacf5_AU17 | AU17 | x_pacf5 | 1st-order difference (velocity) | Rhythm of chin movement velocity; how predictable the speed of the chin movements is from its state at preceding time points. High values indicate regular, rhythmic velocity patterns. |
| diff1x_pacf5_AU01 | AU01 | x_pacf5 | 1st-order difference (velocity) | Rhythm of brow movement velocity; regularity in the speed of inner brow raising and lowering over time. |
| diff2_acf1_AU01 | AU01 | x_acf1 | 2nd-order difference (acceleration) | Smoothness of brow movement acceleration; how closely the acceleration of brow movements at one time point predicts the acceleration at the next. Low values indicate jerky, erratic changes in movement speed. |
| diff2_acf10_AU12 | AU12 | x_acf10 | 2nd-order difference (acceleration) | Rhythm of mouth movement acceleration; overall temporal dependency in the acceleration profile of mouth-related movements across short-to-medium time lags. |
| max_kl_shift_AU12 | AU12 | max_kl_shift | None (raw series) | Distributional stability of mouth movement; the largest abrupt change in the statistical distribution of mouth muscle activation between consecutive time windows. High values indicate sudden shifts in movement behavior. |
| diff1_acf1_AU12 | AU12 | x_acf1 | 1st-order difference (velocity) | Smoothness of mouth movement velocity; how predictable the velocity of lip corner pulling is from one from to the next. |
| lumpiness_AU12 | AU12 | lumpiness | None (raw series) | Variability consistency of mouth movement; how unevenly distributed the variability of mouth muscle activation is over time. High values indicate alternative periods of high and low movement variability. |
| diff2_acf1_AU12 | AU12 | x_acf1 | 2nd-order difference (acceleration) | Smoothness of mouth movement acceleration; temporal predictability of the acceleration profile of mouth movements. |

**S5. Detailed classifier performance.**

Full performance metrics for all four classifiers are presented in Table 1 of the main text. Here we note key patterns across models.

The three non-linear classifiers (Random Forest, C5.0B, SVM) all achieved significant above-chance performance ($p < .01$), while Logistic Regression did not (accuracy = 56.4%, $p = .128$). This convergence across architecturally distinct non-linear models provides robust evidence that the eight selected temporal features carry a genuine discriminative signal. The failure of the linear baseline is consistent with the non-linear decision boundary visualized in Figure 2E, indicating that the bio-behavioral fingerprint of face-swapped deepfakes manifests as a complex, non-linear pattern across multiple temporal features rather than a simple directional shift in any individual feature.

Among the non-linear models, sensitivity-specificity trade-offs varied. C5.0B achieved the highest sensitivity (78.7%) but the lowest specificity among non-linear models (51.1%), suggesting a bias toward classifying videos as fake, likely amplified by the boosting procedure's iterative upweighting of misclassified examples. SVM showed a more balanced profile (sensitivity = 61.7%; specificity = 66.0%) and a ROC-AUC (0.692) comparable to Random Forest (0.694), suggesting similar discriminative capability across thresholds despite slightly lower accuracy. Random Forest's modest performance advantage is likely attributable to ensemble averaging providing greater robustness to noise in the feature space.

**S6. Emotion-matched sensitivity analysis.**

The training set contained a mild imbalance in emotional content (Emotion: n=222, 60%; No Emotion: n=148, 40%). To rule out the possibility that the emotion-dependent performance advantage (Figure 3) was driven by this imbalance, we retrained the Random Forest classifier on an emotion-balanced training set created by downsampling the Emotion class to match the No Emotion class (n=148 each; total n=296). Downsampling was preferred over oversampling as it provides a more conservative test by removing information from the advantaged class. All other pipeline parameters were held constant, and the test set retained its original distribution.

Results confirmed that the emotion-dependent pattern persists after balancing (Table S3). On Emotion videos, the balanced model achieved 67.2% accuracy ($p < .01$) with a ROC-AUC of 0.727 [95% CI: 0.596-0.857]. On No Emotion videos, performance remained near chance (58.3%, $p = .203$; ROC-AUC = 0.546 [95% CI: 0.343-0.749]). The accuracy gap narrowed from 16.9 to 8.9 percentage points, as expected given the reduced number of Emotion training examples, but the qualitative pattern was unchanged: significantly above-chance detection for Emotion videos and chance-level detection for No Emotion videos. This rules out training set imbalance as the driver of the emotion-dependent effect.

**Table S3.** Deepfake detection performance on Emotion and No Emotion test subsets, comparing models trained on imbalanced versus balanced training data.

|  | **Imbalanced Model (222E/148NE)** | | **Balanced Model (222E/222NE)** | |
| --- | --- | --- | --- | --- |
|  | Emotion (n=58) | No Emotion (n=36) | Emotion (n=58) | No Emotion (n=36) |
| ROC-AUC | 0.765 | 0.605 | 0.727 | 0.546 |
| Accuracy | 72.4% | 55.6% | 67.2% | 58.3% |
| P-value | < .001 | .309 | < .01 | .203 |
| Sensitivity | 65.5% | 83.3% | 65.5% | 88.9% |
| Specificity | 79.3% | 27.8% | 69.0% | 27.8% |
| Δ Accuracy | 16.9 pp | | 8.9 pp | |

*Note: Original model trained on full training set (n=370); balanced model trained on emotion-balanced subset (n=296). Both evaluated on the same held-out set.*

**S7. Valence classification.**

To confirm that the results of the balanced valence classification analysis reported in the main text were not driven by the downsampling procedure, we repeated the analysis using the full unbalanced training set (positive: n=144; neutral: n=154; negative: n=72). Results are presented in Table S4. The same directional pattern was observed, with higher classification performance on real compared to fake test videos. Per-class sensitivity patterns were also consistent. Results converged across balanced and unbalanced analyses.

**Table S4.** Valence classification performance on unbalanced real versus fake test subsets.

| Test Subset | ROC-AUC | Accuracy | MCC | P-value | SensPos | SensNeu | SensNeg |
|---|---|---|---|---|---|---|---|
| Real | 0.870 | 0.745 | 0.621 | < .001 | 0.938 | 0.895 | 0.250 |
| Fake | 0.805 | 0.681 | 0.512 | < .001 | 0.812 | 0.842 | 0.250 |
| Δ(Real-Fake) | 0.065 | 0.064 | 0.109 | - | 0.126 | 0.053 | 0 |

**S8. Principal Component Analysis (PCA) as an alternative feature selection method.**

As a sensitivity analysis, PCA was applied to the full set of CMFTS features to explore whether a variance-based representation could provide a more stable feature set. Retaining components explaining 95% of cumulative variance required 51 principal components (Figure S2), indicating that variance in facial dynamic features is widely distributed. The Kaiser criterion (eigenvalues > 1) (10) suggested 25 components, which were used as a more parsimonious representation.

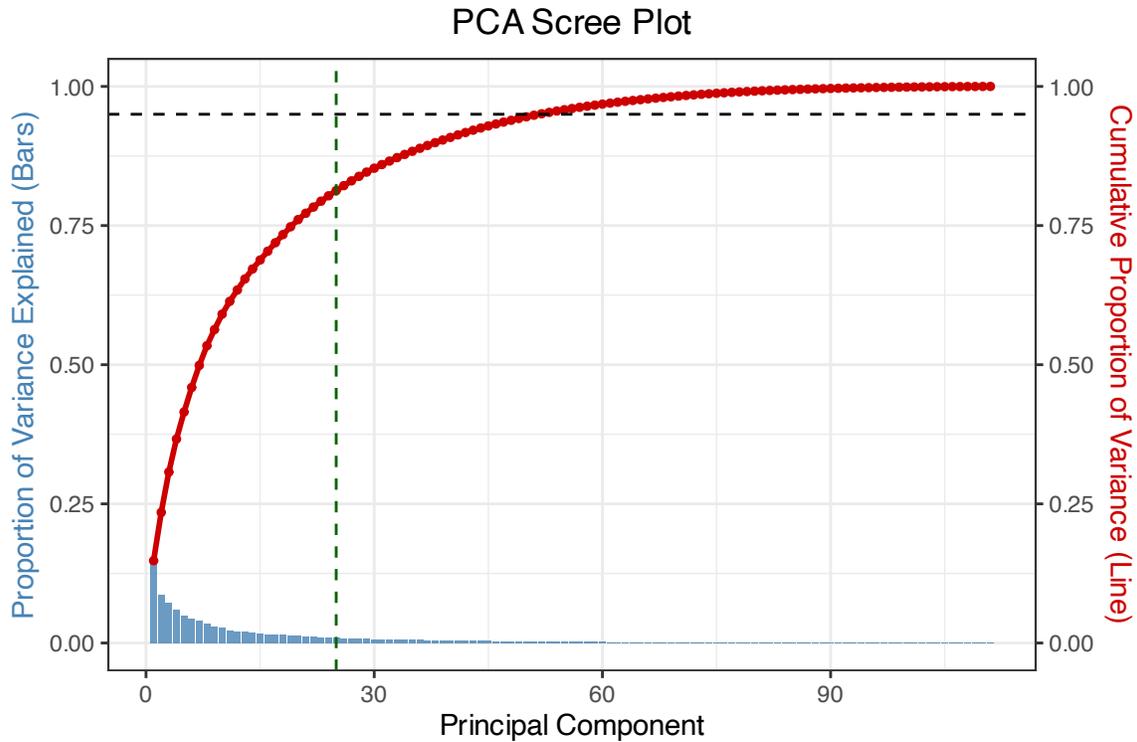

**Fig. S2.** Scree plot showing variance explained by principal components derived from CMFTS features calculated from the three NMF components. Bars indicate proportion of variance explained by each component (left y-axis); red line shows cumulative variance (right y-axis). Dashed horizontal line indicates 95% cumulative variance (51 components); dashed vertical line indicates Kaiser criterion cutoff (25 components).

Classifiers trained on 25 PCA components showed worse performance than the representative AU pipeline (Table S5). The best-performing model was C5.0B (accuracy = 58.5%, $p$ = .061). All remaining models achieving 55.3% accuracy ($p$ = .177), likely due to PCA producing an uninformative space across classifiers. None reached statistical significance.

**Table S5.** Classifier performance using 25 PCs.

| Model | Accuracy [95% CIs] | Kappa | P-value | Sensitivity | Specificity |
|---|---|---|---|---|---|
| Random Forest | **0.553** [0.447-0.656] | 0.106 | 0.177 | 0.489 | 0.617 |
| C5.0B | **0.585** [0.479-0.686] | 0.170 | 0.061 | 0.426 | 0.745 |
| SVM | **0.553** [0.447-0.656] | 0.106 | 0.177 | 0.575 | 0.532 |
| Logistic Regression | **0.553** [0.447-0.656] | 0.106 | 0.177 | 0.511 | 0.596 |

These results suggest that the detection signal is distributed across many subtle temporal characteristics not captured by high-variance dimensions. This validates the Boruta-based feature selection approach used in the main analysis.

**S9. Transition event analysis.**

To investigate whether the predictive signal was concentrated in periods of rapid facial movement, we isolated transition frames defined by AU velocity exceeding a global threshold (mean ± SD of frame-to-frame first-order differences across the three representative AUs, computed on the training set).

The distribution of AU velocities was highly concentrated around zero (mean = 0.0002; Figure S3), with most frames representing minimal movement. Videos containing emotive expressions showed a higher proportion of transition frames (20.86%) than neutral videos (13.63%), with positive valence expressions the highest (25.5%). Critically, real and fake videos had nearly identical proportions (real = 18.13%, fake = 17.80%), indicating that face-swapping does not systematically alter the overall quantity of movement.

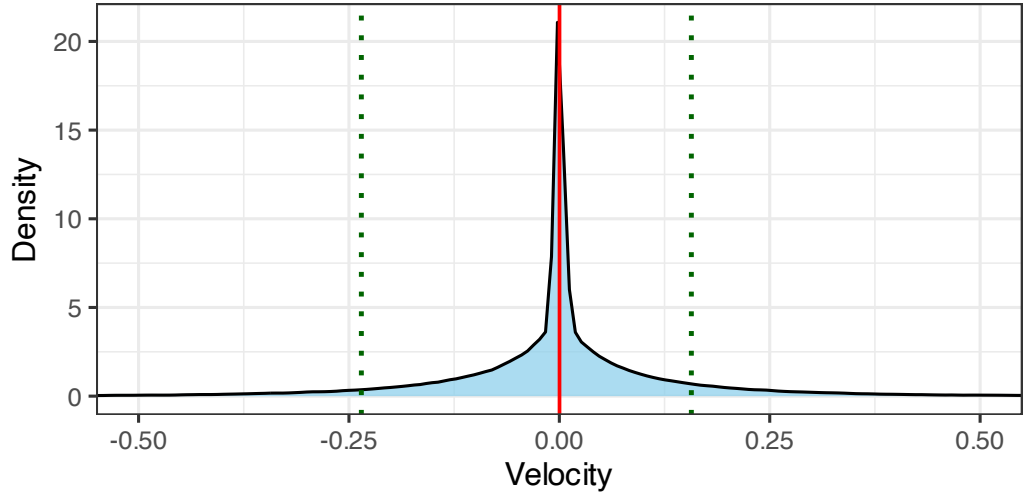

**Fig. S3.** Distribution of frame-to-frame AU velocity across all training videos. Red vertical line indicates mean velocity (0.0002); green dotted lines indicate ± 1 SD thresholds defining transition events.

Contiguous blocks of transition frames were treated as discrete dynamic events, for which summary features were calculated (duration, mean velocity magnitude, maximum velocity, velocity range) and aggregated to the video level. Classification models trained exclusively on these features performed at chance (Table S6), with neither Random Forest (accuracy = 56.4%, $p$ = .128) nor C5.0B (accuracy = 54.2%, $p$ = .235) reaching significance.

**Table S6.** Classifier performance on transition features.

| Model | Accuracy [95% CI] | Kappa | P-value | Sensitivity | Specificity |
|---|---|---|---|---|---|
| Random Forest | **0.564** [0.458-0.666] | 0.128 | 0.128 | 0.553 | 0.575 |
| C5.0B | **0.542** [0.436-0.646] | 0.085 | 0.235 | 0.596 | 0.489 |

These results indicate that the predictive signal is distributed throughout the full temporal structure of facial dynamics rather than concentrated in isolated transition events, consistent with the superior performance of autocorrelation-based features in the representative AU pipeline.

**S10. Human-model convergence.**

**S10.1 Feature selection for human judgments.** A separate Boruta analysis using individual human binary judgments as the prediction target identified 34 features as statistically important. Only four overlapped with the eight features diagnostic for deepfake detection: *diff1_acf1_AU12*, *diff2_acf1_AU12*, *diff2_acf10_AU12*, and *max_kl_shift_AU12* (all mouth movement features). The detection model's two most important features (*diff1x_pacf5_AU17* and *diff1x_pacf5_AU01*) were not confirmed as important for predicting human judgments.

The human-predictive features were more heavily weighted toward entropy-based measures (approximate, sample, and spectral entropy for AU12 and AU17), long-range temporal dependency (Hurst exponent for AU12 and AU17), trend-based statistics (AU01 and AU17), and stationarity measures (unit root tests for AU12).

**S10.2 Predicting human judgments from temporal features.** Random Forest classifiers trained on the 34 human-predictive features were evaluated using two cross-validation strategies. Leave-One-Participant-Out (LOPO) cross-validation achieved 59.7% accuracy [95% CI: 58.1-61.4%] with Kappa = 0.163 ($p < .001$) but exhibited a specificity bias (specificity = 79.5%; sensitivity = 36.3%), suggesting a default tendency to predict that humans would judge videos as real. Leave-One-Stimulus-Out (LOSO) cross-validation achieved 56.9% accuracy [95% CI: 55.3-58.6%] with Kappa = 0.092 ($p < .001$) and a stronger specificity bias (specificity = 84.0%; sensitivity = 24.8%). The low LOSO sensitivity indicates that the features capture participant-level tendencies rather than stimulus-level perceptual cues that generalize across videos.

**S10.3 Extended predictions.** LOPO-predicted human judgments were generated for the broader test set (n=94) and full dataset (n=464) to assess whether human-model correspondence extends beyond the held-out subset (n=40). The LOPO model was used for extension as it achieved higher cross-validated accuracy than the LOSO model, though applying it to unseen stimuli effectively tests stimulus-level generalization.

Predicted human accuracy was 49.5% [95% CI: 39.3-59.6%] on the n=94 test set and 53.8% [95% CI: 49.2-58.3%] on the full dataset, broadly consistent with observed human accuracy of 50.5%. The predicted judgments exhibited a sensitivity bias on the extended datasets, contrasting with the specificity bias during cross-validation on the n=40 set, likely reflecting the interaction between learned participant-level tendencies and different base rates of emotive content in the broader datasets.

No significant alignment was found between predicted human judgments and detection model predictions on either the n=94 test set ($\chi^2(1) = 0.00$, $p = 1.00$, $\varphi = 0.00$). A significant but weak correspondence emerged on the full dataset ($\chi^2(1) = 4.36$, $p = .037$, $\varphi = 0.10$), though this should be interpreted cautiously given that the n=464 set includes training data and the effect size is small.